\documentclass[lettersize,journal]{IEEEtran}
\usepackage{amsmath,amsfonts}
\usepackage{algorithmic}
\usepackage{algorithm}
\usepackage{array}
\usepackage{textcomp}
\usepackage{stfloats}
\usepackage{url}
\usepackage{verbatim}
\usepackage{graphicx}
\usepackage{cite}
\usepackage{indentfirst}
\usepackage{booktabs}
\usepackage{marvosym}
\usepackage{makeidx}
\begin{document}

\title{Deep Vision in Analysis and Recognition of Radar Data: Achievements, Advancements and Challenges}

\author{ Qi Liu \textsuperscript{1},~\IEEEmembership{Senior Member,~IEEE,} Zhiyun Yang \textsuperscript{1}, Ru Ji, Yonghong Zhang, Muhammad Bilal$^*$,~\IEEEmembership{Senior Member,~IEEE}, Xiaodong Liu,~\IEEEmembership{Senior Member,~IEEE,} S Vimal,~\IEEEmembership{Senior Member,~IEEE,} and Xiaolong Xu

        % <-this % stops a space
% \thanks{Corresponding Author: Muammad Bilal (m.billa@ieee.org)}% <-this % stops a space
\thanks{\indent This work was supported by the Key Laboratory Foundation of National Defense Technology under Grant 61424010208, National Natural Science Foundation of China (No. 41911530242 and 41975142), 5150 Spring Specialists (05492018012 and 05762018039), Major Program of the National Social Science Fund of China (Grant No. 17ZDA092), 333 High-Level Talent Cultivation Project of Jiangsu Province (BRA2018332), Royal Society of Edinburgh, UK and China Natural Science Foundation Council (RSE Reference: 62967\_Liu\_2018\_2) under their Joint International Projects funding scheme, and basic Research Programs (Natural Science Foundation) of Jiangsu Province (BK20191398).}
\thanks{\indent Q. Liu is with the Wuxi University, 333 Xishan Avenue, Wuxi 214105, China, and also with the School of Computer and Software, Nanjing University of Information Science and Technology, Nanjing 210044, China (e-mail: qi.liu@nuist.edu.cn).}
\thanks{\indent Z. Yang is with the School of Computer and Software, Nanjing University of Information Science and Technology, Nanjing 210044, China (e-mail: zhiyunyang@nuist.edu.cn).}
\thanks{\indent R. Ji is with the School of Computer and Software, Nanjing University of Information Science and Technology, Nanjing 210044, China (e-mail: 201883290548@nuist.edu.cn).}
\thanks{\indent M. Bilal is with the Division of Computer and Electronic Systems Engineering, Hankuk University of Foreign Studies, South Korea (e-mail: m.bilal@ieee.org).}
\thanks{\indent X. Liu is with the School of Computing, Edinburgh Napier University, Edinburgh EH10 5DT, U.K. (e-mail: x.liu@napier.ac.uk).}
\thanks{\indent S. Vimal is with the Department of Artificial Intelligence \& DS, Ramco Institute of Technology, Tamilnadu, India (e-mail: XXX).}
\thanks{\indent X. Xu is with the School of Computer and Software, Nanjing University of Information Science and Technology, Nanjing 210044, China (e-mail: xlxu@ieee.org).}
\thanks{\indent \textsuperscript{1}Both authors are the first author due to equal contribution to this paper.}
\thanks{\indent Corresponding Author: Muhammad Bilal}}

\markboth{Article accepted for publication in IEEE Systems, Man and Cybernetics Magazine}%
{Shell \MakeLowercase{\textit{et al.}}: A Sample Article Using IEEEtran.cls for IEEE Journals}
% \markboth{Journal of \LaTeX\ Class Files,~Vol.~XX, No.~XX, July~2022}%


\maketitle

\begin{abstract}
% Weather forecasting is regarded as a critical yet challenging task in meteorology. 
Radars are widely used to obtain echo information for effective prediction, such as precipitation nowcasting. In this paper, recent relevant scientific investigation and practical efforts using Deep Learning (DL) models for weather radar data analysis and pattern recognition have been reviewed; particularly, in the fields of beam blockage correction, radar echo extrapolation, and precipitation nowcast. Compared to traditional approaches, present DL methods depict better performance and convenience but suffer from stability and generalization. 
% Further optimization bringing in leading-edge but fine-grained mechanisms and ideas is expected focusing into detailed processes, e.g., blurred radar extrapolation outputs, unclear edges, inconspicuous feature extraction, and so on. 
In addition to recent achievements, the latest advancements and existing challenges are also presented and discussed in this paper, trying to lead to reasonable potentials and trends in this highly-concerned field.
\end{abstract}


\section{Introduction}
\IEEEPARstart{N}{owadays}, radars are widely used to conduct environmental (and social) exploration and analysis, including meteorology, hydrology, traffic monitoring, etc. \cite{ref1, ref2, ref3}. In Particular, weather radars have been extensively used to observe, measure and forecast potential severe convective weather, e.g., thunderstorms, heavy precipitation, etc. \cite{ref4}. The interchange of significant volumes of radar data in a highly fluctuating environment is necessary for further applications in the meteorological area. For example, Internet of weather Radars (IoRs) can be employed for the observation and analysis of high-resolution signals from widespread water particles in the atmosphere \cite{ref5}.

Following the recent development of cutting-edge technologies, such as IoT, Artificial Intelligence (AI) and so on, radar data analytics and relevant services has become highly concerned with interesting outcomes \cite{ref6}. Recent Deep Learning (DL)-based efforts have depicted rapid advancement within various workflow processes of radar echoes, e.g., Blockage Correction (Quality Control), Echo Extrapolation (Time-series Prediction), Nowcasting (Final Production), etc. \cite{ref7,ref8,ref9,ref11,ref20,ref21,ref22,ref23,ref24,ref25,ref26,ref27,ref28,ref29,ref30,ref36,ref37,ref38,ref39,ref40,ref41,ref42,ref43,ref44,ref45,ref46,ref47,ref48,ref49}. However, such achievements above still suffer from radar data without consistent standards and massive use of computational resources, considering data analysis and recognition of weather radars.

In this paper, recent contribution and near-future trend on beam blockage correction, radar echo extrapolation, as well as short-term precipitation forecast are discussed. In Section 2, a brief review of related research on three topics is provided. After that, latest development on the topics is examined, followed by a final conclusion on the weather radar data analysis and recognition.

\section{Related Works}
\subsection{Research Progress of Beam Blockage Correction}
The main research idea of the classic weather radar beam blockage correction method is to manually observe the data rules, design the rules and customize the model, and fill in the adjacent data based on the context. However, because the rules of manual observation have certain limitations, and the deep rules of massive radar data cannot be fully utilized. The effect is not good. In recent years, researchers have conducted in-depth research on the optimization of image completion quality, speed and details with the rapid development of deep learning technology in the field of image completion. These researches have achieved good results.  

\IEEEpubidadjcol
\subsubsection{Classic Beam Blockage Correction Methods}
A classic weather radar beam blockage correction method is a correction method that relies on terrain data, which refers to a Digital Elevation Model (DEM). A dynamic weather radar beam blocking correction method was proposed by Zhang et al from the Chongqing Meteorological Bureau, which is realized by a multiplication factor between the two antenna elevation angles \cite{ref7}. Rainfall radar measurements in mountainous areas were discussed by Andrieu et al \cite{ref8}. When checking weather radar data, special attention should be paid to the influence of terrain and altitude, and a digital terrain model was used for correction. The correction efficiency depends on the accuracy of the radar antenna pointing. The effectiveness of radar rainfall measurement in mountainous areas were studied by comparing with rainfall data, and evaluated the correction of beam blocking and vertical distribution of reflectivity to improve radar measurement accuracy \cite{ref9}.

Another classic weather radar beam blockage correction method is a terrain-independent correction method. A new spatial analysis technique was developed by McRoberts to objectively identify areas where precipitation estimates are contaminated by beam blocking \cite{ref11}. The method requires only long-term precipitation climatology and does not require knowledge of the terrain or the prerequisites of known obstacles.

\subsubsection{Relevant Image Completion Methods}
In recent years, the emergence of deep neural network technology has effectively promoted the development of the field of image completion. According to the type of network architecture, the methods are divided into five categories: Context-Encoder, U-Net, CGAN, DCGAN, and StackGAN.

% Structural loss was proposed, which is a linear combination of pixel reconstruction loss and feature reconstruction loss to further improve the accuracy of the method and is relied on the Context-Encoder framework \cite{ref14}. A multi-scale attention module was added to U-Net to calculate not only the similarity of low-level details of images, but also high-level semantics of images \cite{ref16}. A two-branch network Pluralistic based on the CGAN architecture was proposed \cite{ref17}. One branch is the reconstruction path, and the prior distribution of the missing area information is obtained according to the known area information in the image. The other branch is the generation path, which combines the above information obtained in one branch generates the defective part of the content. Under the StackGAN architecture system, a coarse-to-fine two-stage network architecture GIICA was proposed. The first stage uses reconstruction and the second stage is trained with the reconstruction loss and the adversarial loss. The encoder in the second stage can learn better feature representations than the first stage, and can obtain a more complete scene \cite{ref19}.

Although there have been many achievements in the field of image completion in recent years, there are few relevant researches on this kind of technology in the field of weather radar correction. Judging from the research results in the field of image completion, it has great potential to apply this kind of technology to weather radar block correction, and it is expected to achieve better correction effects.

\subsection{Research Progress of Radar Echo Extrapolation}
At present, numerous deep learning methods are applied to radar echo extrapolation, and these methods can be broadly classified into CNN-based methods, RNN-based methods, and hybrid neural network methods.
\subsubsection{CNN-based Methods}
A U-Net model was proposed for precipitation forecasting based on CNN, which is a well-known encoder-decoder architecture for precipitation nowcasting based on radar data \cite{ref22}. The SmaAt-UNet model was defined. This model is an efficient convolutional neural networks-based on the U-Net architecture equipped with attention modules and depth wise-separable convolutions, which improves the model’s ability to make short-term forecasts with the latest captured information from input data \cite{ref23}. The TRU-NET (Temporal Recurrent U-Net) was brought, which is a model with a novel 2D cross attention mechanism between contiguous convolutional-recurrent layers that improves the modeling of processes defined at multiple spatiotemporal scales \cite{ref24}. A novel architecture based on the core U-Net model named Broad-UNet was adapted, which is able to capture multi-scale information efficiently \cite{ref25}.

The above CNN-based methods possess different advantages, mainly in the ability to capture short-term motion and multidimensional scale information. However, the CNN structure-based approaches lead to spatial location concentration due to the recurrence of each prediction frame, which makes the CNN-based methods relatively weak in capturing long-term motion.
\subsubsection{RNN-based Methods}
Recurrent neural networks are widely used in spatiotemporal sequence prediction to capture features. The ConvLSTM model was proposed and got better extrapolation results than traditional methods \cite{ref26}. The Trajectory GRU (TrajGRU) model was defined, which actively learns the position change of the recursive connections using the subnet output state-to-state connection structure before the state transition \cite{ref27}. A predictive recurrent neural network (PredRNN) was adapted to model spatial representations and temporal changes, extracting both memory space and temporal representations in a stacked RNN structure \cite{ref28}. Subsequently, a novel recursive structure called Causal LSTM was proposed, which is constructed with cascaded dual memory to enhance the ability of PredRNN++ to model short-term dynamics \cite{ref29}. The Memory-In-Memory (MIM) network was presented to optimize original forget gate in original LSTM unit. The network replaces forget gate with two cascaded LSTMs, making learning features captured from spatiotemporal sequences smoother \cite{ref30}.

Compared with traditional neural networks, the RNN-based approaches have the advantage of handling data with arbitrary input and output lengths, while the images predicted by the RNN structure-based model become blurred due to the loss of fine-grained visual appearances. This poses a challenge to our processing of visual representations of radar images.
\subsubsection{Hybrid Neural Network Methods}
In many innovative models, more than just a neural network is used. A Multi-Level Correlation Long Short-Term Memory (MLC-LSTM) model was proposed. The model uses an RNN-structured encoder-predictor and a CNN-structured discriminator to solve the echo evolution problem and the echo prediction ambiguity problem \cite{ref36}. Respectively, the residual convolution LSTM (rcLSTM) and the Generative Adversarial Networks-rcLSTM (GAN-rcLSTM) were presented. The former introduces a residual module to overcome the degeneracy phenomenon of LSTM networks. The latter introduces discriminators to solve the ambiguity problem in long sequence prediction \cite{ref37}. A two-stage extrapolation model based on 3D Convolutional Neural Network (3D-CNN) and Conditional Generative Adversarial Network (CGAN) named ExtGAN was defined. This model can more accurately forecast convective cells that usually lead to severe hazards \cite{ref38}. A new ConvRNN model of energy-based GAN named EBGAN-Forecaster was built. This model effectively alleviates the problems of ambiguity and unrealism and is more stable \cite{ref39}.

Although the models of hybrid neural networks are able to deal with the ambiguity of radar image prediction, the problems of each neural network are reflected in their respective existence. The model based on GAN structure will be unstable when performing training because it is difficult to reach Nash equilibrium.
\subsection{Research Progress of Short-term Precipitation Forecast }
\subsubsection{Traditional Methods}
In a variety of application situations, quantitative precipitation nowcasting (QPN) has become a crucial approach. The motion of precipitation features is tracked from a series of weather radar images, and the precipitation field is then displaced to the near future (minutes to hours) based on that motion, assuming that the strength of the features remains constant.

As an alternative to the trivial case of Eulerian persistence, a series of benchmark processes for quantitative precipitation nowcasting was devised \cite{ref45}. The Pyramid Lucas-Kanade Optical Flow (PLKOF) approach was introduced \cite{ref46}. The capacity to identify big and minor displacement motions utilizing a multi-resolution data structure is a benefit of the PLKOF approach.

Due to excessive prediction smoothing, deep learning nowcasting models suffer from conditional bias, exhibiting worse skill on severe rain rates than Lagrangian persistence models. The concept of model stacking for improving deep learning prediction skills was introduced \cite{ref50}, specifically for intense precipitation regimes. When merging the aggregate with atmospheric circulation features, the suggested strategy doubles the forecasting performance of a deep learning model on severe precipitation.
\subsubsection{Machine Learning Methods}
Deep learning derives low-level picture features on the lowest layers of a hierarchical network and increasingly abstract features on the higher network layers as part of the solution of an optimization problem based on training data, rather than depending on engineering features.

Precipitation nowcasting was defined as a spatiotemporal sequence forecasting problem, ConvLSTM introduced which was a new LSTM extension that outperformed earlier models in a time-series assignment for pictures. Using this as a foundation, a new recurrent structure called Causal LSTM and a Gradient Highway Unit to solve the gradient propagation problem was developed \cite{ref29}.

Convcast, a new precipitation nowcasting architecture that uses satellite data to anticipate diverse short-term precipitation occurrences was proposed \cite{ref48}. Satellite-based precipitation nowcasting is quite important as radar data has limitations of not being available in all regions.

SmaAt-UNet model was proposed \cite{ref23}, which is a convolutional neural network with attention modules and depthwise-separable convolutions based on the well-known UNet architecture. The benefit of the SmaAt-UNet model is that the model parameter size is reduced to a fourth of the original UNet implementation while keeping equivalent performance to the original UNet architecture, allowing precipitation models to be used on small processing units like smartphones.

RainNet, a deep convoutional neural network for radar-based precipitation nowcasting, was presented \cite{ref47}. RainNet had discovered the best level of smoothing for producing a nowcast with a 5-minute lead time. The decrease of spectral power at tiny sizes is also instructive in this regard, since it shows the limits of prediction as a function of geographic scale.

Using mathematical, financial, and neurological metrics, a deep generative model for probabilistic nowcasting of precipitation from radar was described \cite{ref6}. The result shows that generative nowcasting can produce probabilistic forecasts that increase forecast value and operational usability at resolutions and lead periods where other approaches fail.

A radar-based precipitation nowcasting model using an advanced machine learning technique was developed \cite{ref49}, conditional generative adversarial network (cGAN), named Rad-cGAN. This study reveals that Rad-cGAN can be consistently used to precipitation nowcasting with longer lead periods, and that it performs well in areas outside than the originally trained region when employing the transfer learning technique.

As climate change alters weather patterns and the frequency of extreme weather occurrences rises, providing actionable predictions at high geographical and temporal resolutions becomes increasingly crucial. Such forecasts aid in better planning, crisis management, and the minimization of human and material losses. A DL-based infrastructure can deliver forecasts minutes after fresh data is received, allowing them to be completely integrated into a highly responsive prediction service that may better serve the objectives of nowcasting than traditional numerical approaches.

\begin{figure}[h]
\centering
\includegraphics[width=3.5in]{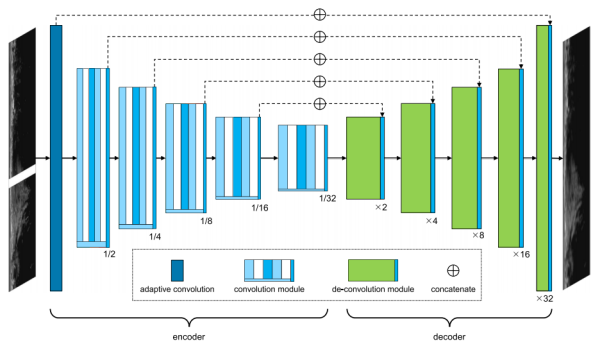}
\caption{Schematic diagram of RC-FCN encoding and decoding network.}
\label{Fig_1}
\end{figure}

\begin{figure}[h]
\centering
\includegraphics[width=3.5in]{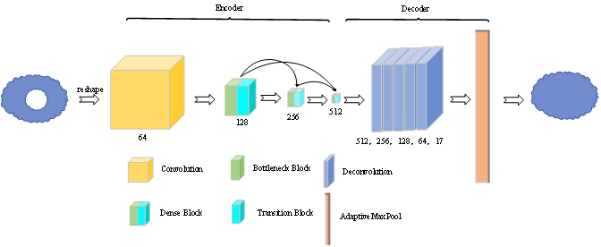}
\caption{Schematic diagram of Dense-FCN encoding and decoding network.}
\label{Fig_2}
\end{figure}
\begin{figure}[h]
\centering
\includegraphics[width=3.5in]{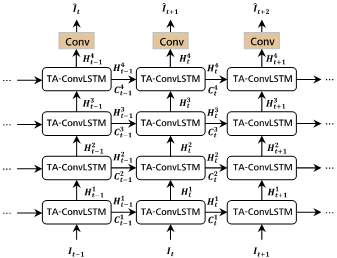}
\caption{The architecture of the radar echo extrapolation network built by stacked four-layer TA-ConvLSTM recurrent cells.}
\label{Fig_3}
\end{figure}
\section{Discussion}
\subsection{Beam blockage correction}
The classic weather radar beam blocking correction methods mainly interpolate and fill through manual observation. These methods cannot take advantage of the deep laws of massive radar data and has limitations. With the continuous development of DL technology in the field of image completion, researchers regard the weather radar beam blockage correction problem as an image completion problem. The introduction of DL technology into the field of weather radar beam blocking correction can make use of the powerful computing power of computers, make full use of the deep laws of massive radar data, and mine the laws that cannot be observed manually for more effective beam correction. An encoder-decoder network (RC-FCN) that combines residual convolution blocks and fully convolutional networks was proposed \cite{ref20}. The specific topology is shown in Figure \ref{Fig_1}. The left side of the network is the encoder network , and the right side is the decoder network. In this experiment, images with missing data and real images were generated manually, and the accuracy of missing region correction was very similar to the real value. To allow DL models to learn more feature maps, some researchers have introduced the idea of  dense connection based on FCN \cite{ref21}. The model contains three dense blocks, and the specific network architecture and process are shown in Figure \ref{Fig_2}. The model harvested more image information and achieved better restoration results. This research also demonstrates the great application prospects and development potential of DL techniques in the field of weather radar beam blockage correction.

\begin{figure}[h]
\centering
\includegraphics[width=3.5in]{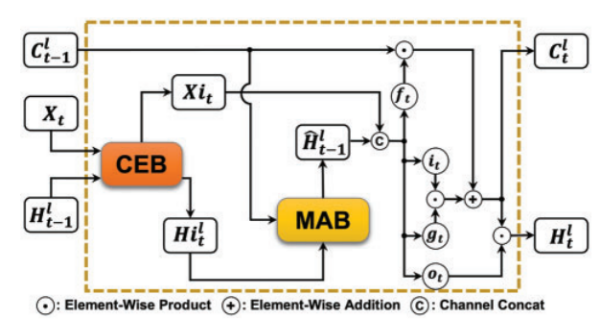}
\caption{The structure of proposed CEMA-LSTM unit.}
\label{Fig_4}
\end{figure}

\subsection{Radar Echo Extrapolation }
\begin{figure}[h]
\centering
\includegraphics[width=3.5in]{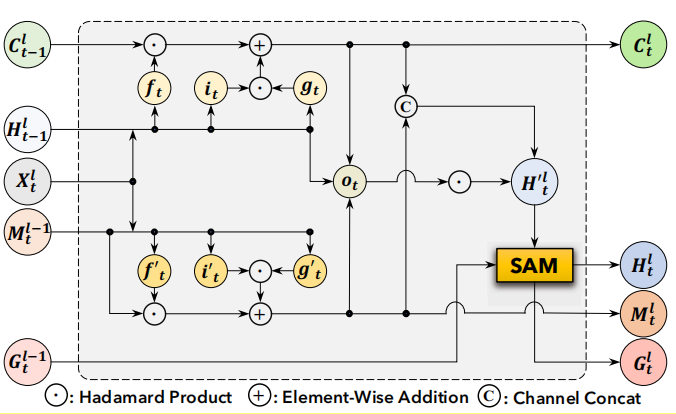}
\caption{General Architecture of PC-Net.}
\label{Fig_5}
\end{figure}

\begin{figure}[h]
\centering
\includegraphics[width=3.5in]{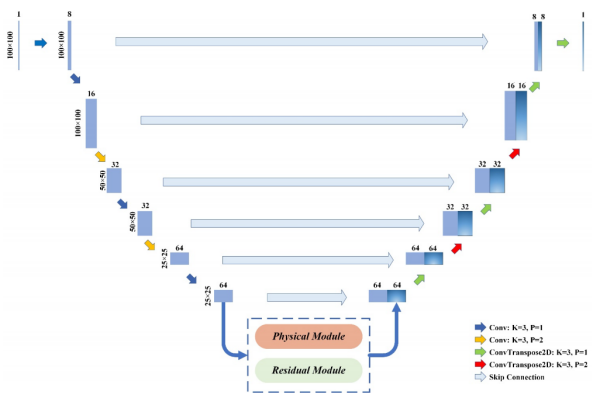}
\caption{The structure of a recurrent unit in the proposed SAST-Net.}
\label{Fig_6}
\end{figure}

\begin{figure*}[!h]
\centering
\includegraphics[width=4.5in]{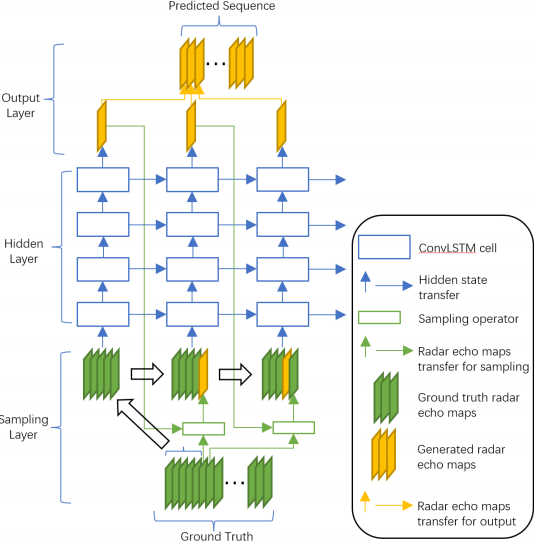}
\caption{The structure of ISS.}
\label{Fig_7}
\end{figure*}

Classical radar echo extrapolation methods mainly use convolution or gated structure in LSTM to capture and store features, thus failing to fully utilize the information in the maps. In recent years, the techniques of DL neural networks in computing have developed rapidly, and more and more researchers are invoking these techniques for radar echo extrapolation. Radar echo extrapolation is considered as a typical spatiotemporal sequence problem, so researchers introduce the concept of spatiotemporal and introduce the stacked structure and multilevel structure into DL. Combined with the current research trends, our group has some results in the field of radar echo extrapolation. The proposed stacked structure models are TA-ConvLSTM \cite{ref40}, CEMA-LSTM \cite{ref41}, SAST-LSTM \cite{ref43}, and their specific network structures are shown in Figure \ref{Fig_3}, Figure \ref{Fig_4}, and Figure \ref{Fig_5}, respectively. The multilevel structure models are PC-Net \cite{ref42}, ISS (An Input Sampling Scheme For RNN-Based Models) \cite{ref44}, and their specific network structures are shown in Figure \ref{Fig_6}, Figure \ref{Fig_7}, respectively. All the above five models can better utilize the information in the radar maps and can mitigate the problem of high intensity echo dissipation with higher prediction accuracy. These five models also face a number of challenges in their future development, such as model instability, edge clarity cannot be guaranteed and the blurring of radar maps. As the technology continues to evolve, the authors will further optimize the models and experimentally propose high-quality models such as GAN-based methods to perform long-term inference tasks.
\begin{figure*}[!h]
\centering
\includegraphics[width=6.5in]{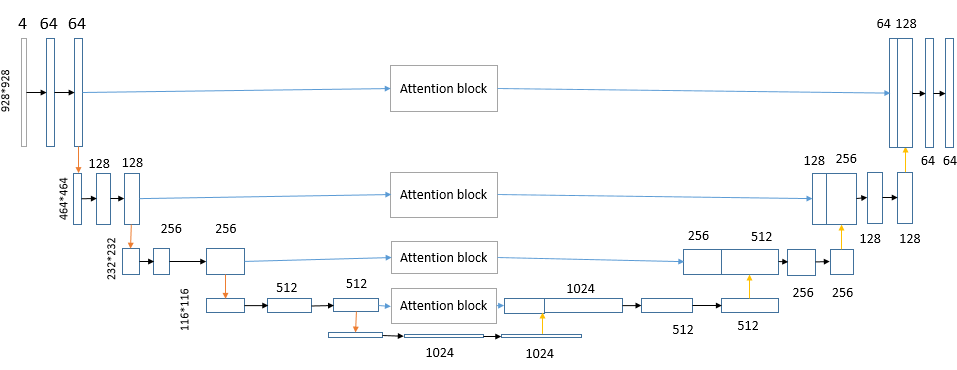}
\caption{Overall structure of SRUNet.}
\label{Fig_8}
\end{figure*}
\subsection{Short-term precipitation forecast}
While classical short-term precipitation forecasting methods are unable to provide actionable predictions at high geographical and temporal resolutions, DL-based methods are able to derive low-level picture features at the bottom layer of hierarchical networks, which can provide predictions in a short time. Short-term precipitation forecasting is defined as a spatiotemporal sequence forecasting problem. On the basis of \cite{ref45,ref47}, a new structure called SRUNet was proposed. The model introduces a self-focus mechanism and its overall structure is shown in Figure \ref{Fig_8}. The model has a great advantage in terms of prediction accuracy. This study shows that DL techniques can serve better for short-term precipitation forecasting compared to traditional numerical methods.

\section{Conclusion And Future Work}
This paper looks at three specific topics in the field of weather radar, i.e., beam blockage correction, radar echo extrapolation, and short-term precipitation forecast. The classic research methods in these three specific research directions and the existing research results combined with DL models have been summarized correspondingly. The latest achievements of DL solutions in these three directions in recent years have been reviewed and discussed.

With the development of cutting-edge DL methods as well as other advancement in the discipline of computer science and atmospheric science, related research effort will certainly meet next improvement and development potentials.

\section*{Acknowledgments}
This work was supported by the Key Laboratory Foundation of National Defense Technology under Grant 61424010208, National Natural Science Foundation of China (No. 41911530242 and 41975142), 5150 Spring Specialists (05492018012 and 05762018039), Major Program of the National Social Science Fund of China (Grant No. 17ZDA092), 333 High-Level Talent Cultivation Project of Jiangsu Province (BRA2018332), Royal Society of Edinburgh, UK and China Natural Science Foundation Council (RSE Reference: 62967\_Liu\_2018\_2) under their Joint International Projects funding scheme, and basic Research Programs (Natural Science Foundation) of Jiangsu Province (BK20191398).

% \newpage

\section*{About The Authors}
\bf{}\vspace{-33pt}
\begin{IEEEbiographynophoto}{Qi Liu}
(M’11, SM’18) received his Ph.D. in Data Telecommunications and Networks from the University of Salford, UK in 2006 and 2010. His research interests include Weather Disaster Monitoring and Warning, Edge-Cloud Collaboration, IoT Continuum, and Smart Grid. His recent research esteem covers but is not limited to Weather Radar Extrapolation, NILM Recognition and Prediction, Distributed Resources Scheduling and Efficiency, Short-term Time-series Data Analytics, etc.
\end{IEEEbiographynophoto}

\vspace{11pt}

\bf{}\vspace{-33pt}
\begin{IEEEbiographynophoto}{Zhiyun Yang}
received his B.E. degree in Computer Science and Technology from Nanjing University of Information Science and Technology, China in 2020. And he is currently studying for a M.S. degree in Computer Science and Technology at Nanjing University of Information Science and Technology, China. His recent research esteem is Weather Radar Echo Extrapolation.
\end{IEEEbiographynophoto}

\vspace{11pt}

\bf{}\vspace{-33pt}
\begin{IEEEbiographynophoto}{Ru Ji}
received her B.E. degree in Computer Science and Technology from Nanjing University of Information Science and Technology, China in 2022. And She is about to start studying for a M.S. degree in Electronic information at Nanjing University of Information Science and Technology. Her main research field is Weather Radar Echo Extrapolation.
\end{IEEEbiographynophoto}

\vspace{11pt}

\bf{}\vspace{-33pt}
\begin{IEEEbiographynophoto}{Yonghong Zhang}
received his Ph.D. degree from Shanghai Jiao Tong University, China in 2005. His research interests include meteorological disaster monitoring and warning, edge cloud cooperation, IoT continuum, and smart grid. His recent research interests include but are not limited to pattern recognition and intelligent systems, remote sensing big data analysis and deep learning, intelligent equipment and IoT system integration, higher education teaching management and and research work.
\end{IEEEbiographynophoto}

\vspace{11pt}

\bf{}\vspace{-33pt}
\begin{IEEEbiographynophoto}{Muhammad Bilal}(M’16, SM’20)
received the Ph.D. degree in information and communication network engineering from the School of Electronics and Telecommunications Research Institute (ETRI), Korea University of Science and Technology, in 2017.  From 2017 to 2018, he was with Korea University, where he was a Postdoctoral Research Fellow with the Smart Quantum Communication Center. In 2018, he joined Hankuk University of Foreign Studies, South Korea, where he is currently working as an Associate Professor with the Division of Computer and Electronic Systems Engineering. His research interests include design and analysis of network protocols, network architecture, network security, IoT, named data networking, Blockchain, cryptology, and future Internet.  
\end{IEEEbiographynophoto}

\vspace{11pt}
\bf{}\vspace{-33pt}
\begin{IEEEbiographynophoto}{Xiaodong Liu}
(M’00, SM’17) received his PhD in Computer Science from De Montfort University and joined Napier in 1999. He is a Reader and is currently leading the Software Systems research group in the IIDI, Edinburgh Napier University. He was the director of Centre for Information \& Software Systems. He is an active researcher in software engineering with internationally excellent reputation and leading expertise in context-aware adaptive services, service evolution, mobile clouds, pervasive computing, software reuse, and green software engineering.
\end{IEEEbiographynophoto}

\vspace{11pt}

\bf{}\vspace{-33pt}
\begin{IEEEbiographynophoto}{S Vimal} is with the Department of Artificial Intelligence \& DS, Ramco Institute of Technology, Tamilnadu, India. He holds a Ph.D. in Information and Communication Engineering from Anna University Chennai and he received Masters Degree from Anna University Coimbatore. His areas of interest include Game Modelling, Artiﬁcial Intelligence, Cognitive radio networks, Network Security, Machine Learning and Big data Analytics. He is a Senior Member of IEEE and holds membership in various professional bodies. 
\end{IEEEbiographynophoto}

\vspace{11pt}

\bf{}\vspace{-33pt}
\begin{IEEEbiographynophoto}{Xiaolong Xu}
received his Ph.D. degree from Nanjing University, China in 2016. He is full Professor at Nanjing University of Information Science and Technology. His research interests include but are not limited to Cloud Computing, Big Data, Edge Computing, and Deep Learning.
\end{IEEEbiographynophoto}

\vfill
\end{document}